\def\year{2021}\relax
\documentclass[letterpaper]{article} 
\usepackage{aaai21}  
\usepackage{times}  
\usepackage{helvet} 
\usepackage{courier}  
\usepackage[hyphens]{url}  
\usepackage{graphicx} 
\usepackage{comment}
\usepackage{color}
\usepackage{natbib}  
\usepackage{caption} 
\usepackage{mathtools}
\usepackage{amsmath}
\usepackage{amssymb}
\usepackage{multirow}
\usepackage[ruled,noend]{algorithm2e}

\usepackage[autostyle]{csquotes}
\usepackage[utf8]{inputenc}
\usepackage{color}
\usepackage{soul}
\usepackage{xcolor}

\author {
    Apoorva Upadhyaya,
    Marco Fisichella,
    Wolfgang Nejdl \\
}
\affiliations {
    L3S Research Center, Leibniz Universität Hannover, Hannover, Germany \\
    upadhyaya@l3s.de, mfisichella@l3s.de, nejdl@l3s.de
}

\title{A Multi-task Model for Sentiment Aided Stance Detection of \\Climate Change Tweets}
\begin{document}

\maketitle

\begin{abstract}
Climate change has become one of the biggest challenges of our time. Social media platforms such as Twitter play an important role in raising public awareness and spreading knowledge about the dangers of the current climate crisis. With the increasing number of campaigns and communication about climate change through social media, the information could create more awareness and reach the general public and policy makers. However, these Twitter communications lead to polarization of beliefs, opinion-dominated ideologies, and often a split into two communities of climate change deniers and believers. In this paper, we propose a framework that helps identify denier statements on Twitter and thus classifies the stance of the tweet into one of the two attitudes towards climate change (denier/believer). The sentimental aspects of Twitter data on climate change are deeply rooted in general public attitudes toward climate change. Therefore, our work focuses on learning two closely related tasks: Stance Detection and Sentiment Analysis of climate change tweets. We propose a multi-task framework that performs stance detection (primary task) and sentiment analysis (auxiliary task) simultaneously. The proposed model incorporates the feature-specific and shared-specific attention frameworks to fuse multiple features and learn the generalized features for both tasks. The experimental results show that the proposed framework increases the performance of the primary task, i.e., stance detection by benefiting from the auxiliary task, i.e., sentiment analysis compared to its uni-modal and single-task variants.
\end{abstract}
\section{Introduction}\label{intro}

Climate change is the burning crisis of our time, and it is happening even faster than we thought. A recent article on the BBC News website\footnote{https://www.bbc.com/news/science-environment-60525591} states that many of the effects of global warming are now simply ``irreversible" according to the latest assessment and that more than $40$\% of the world's population is ``at high risk" from climate.
In fact, according to a report by the Intergovernmental Panel on Climate Change (IPCC), it is very likely that climate change is caused by man-made activities \cite{myhre2013climate}. Despite the scientific consensus on the causes and main impacts of climate change, it remains a controversial topic in public discourse. Therefore, understanding public perceptions plays a critical role in addressing climate change by increasing the public's willingness to accept appropriate action on climate change \cite{shi2015public}.
\par\noindent Recently, social media platforms such as Twitter have played an important role in raising public awareness of the current climate crisis and influencing public attitudes toward climate change \cite{lineman2015talking}. Twitter conversations, however, are often influenced by the polarization of beliefs and, in the case of climate change, are divided into two competing groups, one that believes in climate change (Believers) and a second that is skeptical or denies that climate change is occurring (Disbelievers) \cite{jang2015polarized, cann2021ideological}. The article published on the Euronews website\footnote{https://www.euronews.com/green/2021/11/18/climate-misinformation-is-getting-more-sophisticated-and-experts-say-cop26-progress-could-} after the COP26 conference revealed that scientists have found that climate change deniers are not only skeptical about climate change, but have also led to the problem of delaying climate change by either shifting responsibility or eventually capitulating - the idea that it is not possible to prevent climate change, which often leads to the spread of misinformation \cite{zhou2021confirmation}. Therefore, it is important for government agencies, researchers, and technology companies to monitor such content on social media to identify and intervene in tweets from climate change deniers, which will help combat climate misinformation. This has motivated us to identify such content and understand public attitudes toward climate change by performing the important task of stance detection.
\par \noindent Stance detection is the task of automatically identifying the author's point of view in relation to the target object (for/against/neutral). It has been used to identify social attitudes toward pressing issues (e.g., covid19 vaccination \cite{argyris2021using}, climate change, abortion, feminism \cite{mohammad2016semeval}). In our study, we focus on climate change as a target and perform statement-level stance detection. The goal is to predict the attitude expressed in a single tweet, where the stance is for (believers) or against climate change (deniers). 

\par \noindent Numerous works have classified tweets into favor or against the target using the SemEval 2016 benchmark dataset. The dataset contains $5$ target topics, including climate change, with $29$ denier and $335$ believer tweets \cite{li2019multi,wang2020neural}. However, due to the small number of climate-specific tweets, these techniques do not focus on understanding the specific characteristics of climate change deniers and believers. This is because climate change deniers not only deny climate change, but also disagree with the solutions to combat climate change, which often leads to public negligence \cite{zhou2021confirmation}. Other works that deal with the classification of tweets on climate change lack a suitable architecture that efficiently performs the classification task \cite{kabaghe66classifying}. Therefore, in our work, we use a multi-tasking approach that uses different attention frameworks to classify the attitude of the climate change tweet into one of the two polarized classes (deniers/believers) to identify the denier statements on Twitter.
\par \noindent Sentiment analysis has helped in many other tasks, such as detecting hate speech and sarcasm in a multi-task architecture \cite{majumder2019sentiment}. Studies on climate change have also justified the role of sentiment in conversations about climate change, either by assessing sentiment in tweets from climate change deniers or by examining the emotional impact of climate change data on Twitter \cite{dahal2019topic,el2021novel}. Therefore, in our study, we leverage the sentiment analysis task to decipher the attitude of the tweets.
\par \noindent In addition, we use multiple inputs, i.e., combining tweet text and topic representations, to build a reliable classification model that helps identify the sentiment of the tweeter and determine the correct tweet attitude. Topic representations have helped in detecting fake news by providing more discriminatory power to the model used \cite{DBLP:conf/aaai/GautamVM21}. In our study, topic embedding provides a global context to a single tweet and thus more information and can circumvent the drawback of the short length of the tweet text to efficiently train the proposed system.
\par\noindent Our work focuses on learning two closely related tasks, stance detection and sentiment analysis of tweets on climate change. Stance detection is our main task, which is supported by sentiment analysis as an auxiliary task. We propose a multi-task framework that incorporates feature-specific and shared-specific attention frameworks to fuse multiple features and learn the features for both tasks. Our proposed approach is useful for government agencies and technology companies to detect the attitude of posts (deniers/believers) and curb the spread of such content that denies climate change and is false or misleading to combat climate misinformation.
\par\noindent We summarise the contributions of our work as follows:
\textbf{(i)} We create a new dataset\footnote{The dataset and code are available at the repository:  https://github.com/apoorva-upadhyaya/Climate-Change-Tweets.} for the climate change domain consisting of tweets with the stance and sentiment labels which is beneficial for the research community. \textbf{(ii)} We illustrate the importance to consider the sentiment associated with the tweet while categorising the stance of tweet into favor(believers) or against(deniers) climate change. \textbf{(iii)} We propose a multi-task framework that jointly performs the stance detection (primary) and sentiment analysis (auxiliary) tasks. We integrate feature-specific and shared-specific attention frameworks to integrate information across multiple features and shared tasks to learn features that optimise task performance. Experimental results indicate that the proposed framework increases performance of the primary task, i.e., stance detection by benefiting from the auxiliary task, i.e., sentiment analysis compared to its uni-modal and single-task variants.

\section{{Related Works}}\label{related_works}

Climate change has become one of the greatest challenges of our time. 
Several works explore the role of psychology in climate change by examining the impact of climate change on people's well-being and perceptions \cite{clayton2020climate}. Engaging the public is an important part of addressing climate change. Therefore, social media platforms like Twitter allow anyone to explore and report public viewpoints on the complex issue of climate change \cite{lineman2015talking,dahal2019topic}. 
However, debates and discussions on Twitter about climate change are widely associated with increasing polarization and are often divided into climate change believers and deniers \cite{jang2015polarized}. In order to identify and understand public attitudes toward climate change, the task of identifying attitudes plays an important role.

\subsection{Stance Detection}
Stance detection is about classifying the attitude that the author expresses towards a target object. The author may support the target object, reject it, or have a neutral stance. 
In our work, we focus on the climate change target, where opinions are either for climate change (believers) or against climate change (deniers). There are several climate change specific studies where the goal is to predict a user's attitude \cite{chen2019detecting,tyagi2020polarizing}. It has been found that multiple tweets from the same user can have different stance classes. In order not to miss the denier attitude of a single tweet that could interfere with the implementation of climate change policies, we focus on detecting statement-level stance detection, where the goal is to predict the stance described in a single tweet.
\par \noindent The stance detection of tweets has been studied in a variety of work on the popular SemEVAL 2016 dataset, which includes the $5$ targets including climate change (with $364$ climate change tweets) \cite{vychegzhanin2021new,wang2020neural}. However, in these previous studies, little attention was paid to understanding the characteristics of climate change denier and believer tweets in particular (only $29$ climate change denier tweets in the SemEVAL 2016 dataset). Recently, \cite{luo2020detecting} published a stance-annotated dataset on global warming and proposed an opinion framing task to explore the discourse used in the global warming debate. One of the papers \cite{kabaghe66classifying} classified tweets into three classes based on their attitude toward climate change: $-1$ (negative belief), $0$ (neutral belief), and $1$ (positive belief). However, the lack of an architectural framework led us to propose an efficient model that can efficiently classify a tweet's attitude toward climate change into one of two categories (deniers/believers) in real-time, while taking advantage of the other task.

\subsection{Sentiment Analysis}\label{rw_sent}
Some of the works on stance recognition have emphasized the importance of sentiment \cite{wang2020neural}, while some have cited the orthogonal relationship between stance and sentiment of the statement \cite{sen2020reliability}. However, several works have focused on the sentimental aspects of climate change conversations and justified their role in climate change \cite{cody2015climate,jiang2017comparing}. One recent work \cite{el2021novel} proposes a real-time framework that uses sentiment and emotion analysis to provide meaningful insights into public opinion, and tested the model with tweets posted by Greta Thunberg and her followers on climate change. These studies motivated us to investigate the role of sentiment in classifying tweets on climate change.

\subsection{Multi-Task Learning}
\par \noindent Multi-task learning (MTL) \cite{caruana1997multitask} is a learning paradigm that aims to learn multiple related tasks together in the hope of improving generalization performance for all tasks at hand. 
In our work, we focus on learning stance detection task with the help of sentiment analysis in a multi-tasking system. The study by \cite{ li2019multi} uses sentiment to predict the stance through a multi-task learning model. Another work by \cite{chauhan2019attention} uses sentiment as an auxiliary task to predict attitude. However, in our work, multiple features in the form of text and topic words are used to separate the task-dependent and independent feature spaces and perform both tasks simultaneously by using attention frameworks to focus on the most important feature representations and discarding the useless shared features that may affect the performance of both tasks.
\section{Dataset}\label{dataset}
In this section, we describe the Twitter datasets we use for our experiments. We consider both publicly available tweets about climate change and live tweets about climate change that we collect using hashtags from the previous literature. We first discuss data collection strategies, followed by data labeling and data pre-processing techniques. The \textit{qualitative aspect} and the \textit{temporal analysis} of the dataset is described in the Supplementary.
\subsection{Data Collection Method}\label{data_collect}

Previous works have used hashtags to identify the stances of different groups on social media, creating a method of tagging content by topic \cite{misra-etal-2016-nlds}, we also use this explicit annotation quality of hashtags to collect a larger dataset. First, we select the hashtags for deniers and believers \cite{tyagi2020polarizing} from the previous literature (see row $1$ of the table \ref{tab:seedhash}). We then consider the two publicly available Twitter datasets on climate change, (i) tweet IDs collected from September 21, 2017 to May 17, 2019 using a set of climate change keywords available in the Harvard Dataverse \cite{DVN/5QCCUU_2019}, and (ii) tweet IDs used in the work \cite{samantray2019credibility} collected from 2007 to 2019. We start by retrieving the tweet objects from these publicly available tweet IDs using the Tweepy API\footnote{https://docs.tweepy.org/en/stable/ \label{tweepy}}. We analyzed the hashtags used in the collected tweets and found that the hashtags mentioned in row $2$ of the table \ref{tab:seedhash} are the most frequently used and co-occur with the denier and believer seed hashtags. We draw a random sample of $1000$ tweets containing these most frequently used hashtags separately for both categories and identify $98\%$ tweets as deniers and $99\%$ tweets as believers. We conclude that the final set of seed hashtags (see table \ref{tab:seedhash}) can be used to identify tweets from deniers and believers. 
Because the publicly available datasets contain tweets with a large time span that reflect Twitter trends and topics related to climate change and cover a wide range of audiences, they helped us identify additional relevant hashtags associated with climate change deniers and believers and enriched us with a final set of seed hashtags that can be used for further data collection related to climate change deniers and believers. We then select the unique tweets (excluding retweets) from the collected data that contain either the denier or believer hashtags. We obtain a total of $5,682$ denier and $32,111$ believer tweets after the filtering process based on seed hashtags. 
The collected dataset appears to be relatively small, with a smaller number of denier tweets, so we collect the real-time tweets from July 28 to December 26, 2021 using the live-streaming Tweepy API with the final set of seed hashtags. The number of tweets filtered out as deniers and believers according to the seed hashtags from various sources are listed in the table \ref{tab:data}.

\begin{table}[]
\centering
\scalebox{0.72}{
\begin{tabular}{|l|l|l|}
\hline
\textbf{Sources} & \textbf{Denier Hashtags} & \textbf{Believer Hashtags} \\ \hline
\textbf{Tyagi et al. 2020a} & \begin{tabular}[c]{@{}l@{}}ClimateHoax, YellowVests, \\ Qanon\end{tabular} &\begin{tabular}[c]{@{}l@{}}ClimateChangeIsReal, \\ClimateActionNow,\\ FactsMatter, \\ScienceMatters, \\ScienceIsReal\end{tabular}  \\ \hline
\textbf{\begin{tabular}[c]{@{}l@{}}Most Used \& Co-occur \\from public data \end{tabular}}  & \begin{tabular}[c]{@{}l@{}}GlobalWarmingHoax, \\ClimateChangeHoax,\\ ClimateDenial,ClimateHoax\end{tabular} &\begin{tabular}[c]{@{}l@{}}SaveClimate,\\ ActOnClimate\end{tabular}  \\  \hline
\end{tabular}
}
  \setlength{\belowcaptionskip}{-5pt}
\caption{Denier \& Believer Seed Hashtags}
\label{tab:seedhash}
\end{table}

\begin{table}[]
\centering
\scalebox{0.70}{
\begin{tabular}{|l|l|*{2}{c|}*{2}{c|}}
\hline
\textbf{Dataset} & \textbf{Total Tweets}  & \multicolumn{2}{c|}{\textbf{As per seed Hashtag}} &\multicolumn{2}{c|}{\textbf{\begin{tabular}[c]{@{}l@{}}As per label prop. \\algo (out of \\seed hashtag)\end{tabular}}} \\ \hline
\textbf{-} & \textbf{-}  &\textbf{Denier} &\textbf{Believer} &\textbf{Denier} &\textbf{Believer}\\ \hline
\textbf{{{\begin{tabular}[c]{@{}l@{}}Harvard Data\\ (Littman and\\ Wrubel 2019)\end{tabular}}}} &1322969 &1595 &20886 &1042 &18421\\ \hline
\textbf{\begin{tabular}[c]{@{}l@{}}Credibility Data\\ (Samantray \\and Pin 2019)\end{tabular}} &9672907 &4087 &11225 &2883 &7735 \\ \hline
\textbf{Live tweets} &5711743 &10594 &39787 &9200 &34274 \\ \hline
\textbf{Total tweets} &- &16276 &71898 &13125 &60430 \\ \hline
\end{tabular}
}
  \setlength{\belowcaptionskip}{-5pt}
\caption{Dataset Statistics for Stance Detection Task}
\label{tab:data}
\end{table}

{\SetAlgoNoLine
\begin{algorithm}
\SetAlFnt{\small}
    \SetKwInOut{Input}{input}
    \Input{Graph $G$ with nodes $n$ and edges $e$ with $e_{ij}$ as edge weight between $i \in n$ and $j \in n$}
        \textbf{initialize} $\gamma$=$50/100$ and $i$=0 \;
    \For{$each$  $n$}{
        \ define $l$ = integer(i/$\gamma$); $i$+=$1$ \;
    \For{$each$  $n$}{
    \If{$n$ $not$ $labeled$}{
     \ \textbf{compute} $t$ = neighbors of $n$ \;
     \ \textbf{compute} $t_l$ = labeled neighbors of $n$ \;
     \If{$t_l + l$ $\geq$ $t$}{
     \ \textbf{initialize} $score,c$ \;
     \For{$each$ $t_i \in t$}{
     \ $score+=$ label $t_i * e_{nt_{i}}$; $c$+=$e_{nt_{i}}$;
     }
     }
     }
      }
    }
\caption{Label Propagation Algorithm}
\label{label_algo}

\end{algorithm}
}

\subsection{Data Annotation}\label{data_annotate}
\subsubsection{Stance Detection}
To validate the stances of the collected tweets provided by the hashtag self-annotation technique, we run a variant of label propagation algorithm \cite{tyagi2020polarizing,tyagi2020computational}, which transfers the labels from the seed hashtags to other hashtags (refer Algorithm \ref{label_algo}). The authors who provided the label propagation algorithm claim that their approach is similar to various other works \cite{weber2013secular,garimella2018quantifying}. First, we weight the seed hashtags of believers by $+1$ and those of deniers by $-1$ (as in table \ref{tab:seedhash}). We create a weighted hashtag$*$hashtag co-occurrence graph with all hashtags present in the data, where each node represents a hashtag and an edge is created between the hashtags that occur in the same tweet, where the weight of the edge is proportional to the frequency of their co-occurrence. The weights of the seed hashtags are then transferred to other hashtags as specified in the algorithm \ref{label_algo}. The hashtag scores (believer hashtag = $+1$, denier hashtag = $-1$) are then arithmetically summed for all hashtags that occur in each tweet and then averaged. The final score is then used to classify tweets into deniers ( score $< 0$) or believers ( score $> 0$). In total, we found a set of $13,125$ denier and $60,430$ believer tweets (refer table \ref{tab:data}). Three trained annotators drew a random sample of $1000$ tweets from both categories and manually annotated them. To determine the consistency between the ratings of the annotators, we use the Fleiss-Kappa \cite{spitzer1967quantification} measure and achieve an agreement score of $0.84$, indicating that the annotations are of good quality. We consider the manually annotated tweets as the ground truth and compared them with the annotations found after the label propagation algorithm. We found that $98.40\%$ of denier tweets belong to the denier category and $99.6\%$ of believer tweets belong to the believer category. Therefore, to save time and cost, we consider the labels generated by the label propagation algorithm as the final labels for denier and believer tweets. The steps for data collection and data annotation for the stance detection task are briefly described in Fig. 1 in the Supplementary.

\subsubsection{Sentiment Analysis}
We leverage weak supervision approach to annotate tweets for sentiment analysis. Similar to previous work \cite{singh2021multitask}, we use three sentiment classifiers to generate sentiment labels for each preprocessed text of the tweet, namely (i) VADER \cite{hutto2014vader}: a popular lexicon and rule-based sentiment analysis tool that relies on a dictionary to generate sentiment scores, (ii) TextBlob\footnote{https://textblob.readthedocs.io/en/dev/}: a Python-based library that provides an API for handling common NLP tasks such as sentiment analysis, POS tagging, etc., and (iii) NLTK\footnote{
https://www.nltk.org/index.html \label{nltk}}: a Python bundle that provides a collection of NLP algorithms such as sentiment analysis, NER, etc. This results in $3$ labels (\textit{positive, negative, neutral}) per tweet, from which a single label is finally selected as the sentiment expressed by the tweet based on the majority voting based ensemble method. The data statistics is mentioned in table \ref{tab:data1}. Three trained annotators manually evaluated the labels for $1000$ randomly selected tweets and obtained an inter-annotator agreement score of $0.81$ using the Fleiss-Kappa measure. We consider the final annotations generated after inter-annotator agreement as the ground truth and compared them with the annotations generated using the weak supervision approach and found an accuracy of $97.6\%$. To save time and cost, we consider the annotations generated using the weak supervision approach for the sentiment analysis task.

\begin{table}
\centering
\scalebox{1}{
\begin{tabular}{|l|l|l|l|}
\hline
\textbf{Tweets} & \textbf{Negative} & \textbf{Positive} & \textbf{Neutral} \\ \hline
{Denier} &60.2\% &18.2\% &21.6\% \\ \hline
{Believer} &24.7\% &46.2\% &29.1\% \\ \hline
\end{tabular}
}
 \setlength{\abovecaptionskip}{2pt}
  \setlength{\belowcaptionskip}{-5pt}
\caption{Data Statistics for Sentiment Task}
\label{tab:data1}
\end{table}

\subsection{Data pre-processing} \label{data_preprocess}
Data pre-processing is important because raw tweets without pre-processing are very unstructured and contain redundant and often problematic information that affects the performance of the model training and classification task.
\subsubsection{Text}  We remove mentions, URLs, punctuation, spaces, and unwanted characters such as RT (retweet), CC (carbon copy), and stopwords from the tweet text. We use ekphrasis \cite{baziotis2017datastories} to extract hashtags by segmenting long strings into their individual words. For further text processing, we use the Python toolkit NLTK. The NLTK-based tokenizer is used to tokenize tweets. All words are converted to lowercase letters. Then, we reduce the inflected words by applying the NLTK Wordnet lemmatizer, and then apply PorterStemmer for stemming.
\subsubsection{Topic} We first remove the seed hashtags used for data collection, otherwise topics created can be biased towards the hashtags. The tweet text is pre-processed using the procedure described above. In this study, we use BERTopic modeling, which uses transformer-based embeddings to create easily interpretable topics and their distributions \cite{grootendorst2020bertopic}. This modeling technique has recently gained popularity and provided promising results in previous studies \cite{anwar2021analyzing}, therefore we focus on using BERTopic that detect semantic similarity and integrate topics with pre-trained contextual representations. The tweet text is then fed into the BERTopic library with the \textit{calculate\_probabilities=True}, which creates topics from the data and assigns a probability score to each created topic for each tweet sample in the data. We select the m-most similar topics for each tweet sample, where each topic is represented by the top `p' topic words.
\begin{table}
\centering
\scalebox{0.70}{
\begin{tabular}{|l|l|l|}
\hline
\textbf{Category} &\textbf{Tweet} &\textbf{Topic Words}  \\ \hline
{Denier} &\begin{tabular}[c]{@{}l@{}}CO2 is greening the planet \\ and restoring the rainforest.\\ Its almost like the planet is \\able to self regulate \#ClimateHoax .\end{tabular} &\begin{tabular}[c]{@{}l@{}} nonsense, hoax, science,\\ denial, destroy, ridiculous,\\ myth, planet, hypocrisy...\end{tabular}  \\ \hline
{Believer} &\begin{tabular}[c]{@{}l@{}}Great format and read \#climate\\ \#ClimateActionNow \end{tabular} &\begin{tabular}[c]{@{}l@{}} warm, hot, earth, real,\\ emergency, possible, crisis,\\ urgent, sun, warming,...\end{tabular}  \\ \hline
\end{tabular}
}
\caption{Examples with stance, tweet text and topic words}
\label{tabletopic}
\end{table}
\par \noindent \textit{\textbf{{Role of Topic words as Feature}}} In table \ref{tabletopic}, we present two samples from the dataset that illustrate the importance of considering topic words along with the tweet text for the analysis tasks. It can be observed that the tweet text alone is not helpful in identifying the stance of the tweet in the given samples. However, the addition of topic words gives the tweet more context and information that helps in correctly predicting the denier or believer stance of the tweet. These examples show that the presence of complementary information in the form of topic words aids the process of stance detection.

\section{Methodology}\label{Methodology}
In this section, we outline the working of our proposed multi-task model for the stance detection (primary task) and the sentiment analysis (auxiliary task). The proposed model consists of the following  components : \textit{Feature Extraction, Attention Framework, and Classification Layer}. 
The key factor of multi-task learning is the sharing of features across the tasks. Therefore, we first describe the two variants of the model depending on the framework of multi-task learning and then explain the model components in detail.
\subsection{Variants of the Proposed Model}
\begin{figure}
\centering
  \includegraphics[width=0.85\linewidth]{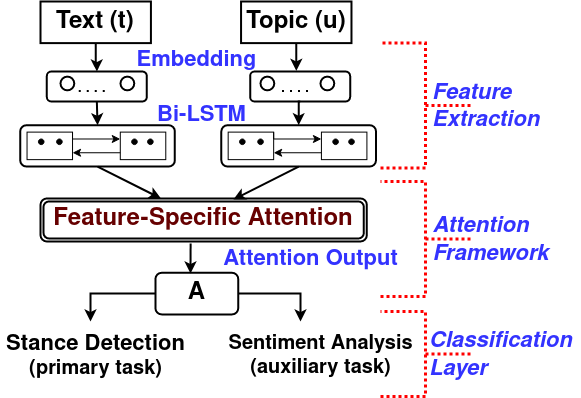}
  \caption{Architectural diagram of the Shared-Only Multi-Task (SO-MT) Framework.}
  \label{shared_only}
  \vspace{-0.3cm}
\end{figure}

We depict the following $2$ variants of the multi-task model depending on the sharing of features across the tasks: (i)\textit{\textbf{Shared-Only Multi-Task Model (SO-MT)}} : In the SO-MT model, we use single shared layers for the feature extractor and the attention framework to extract features for all tasks, as shown in Fig. \ref{shared_only}. The single shared output of the attention framework is then used as an input to the classification layer, which produces separate outputs for the stance and sentiment tasks. This model focuses on the task-invariant features and ignores the fact that some features are task-dependent.
\par
\begin{figure*}
\begin{minipage}{.5\linewidth}
\includegraphics[width=1.02\linewidth]{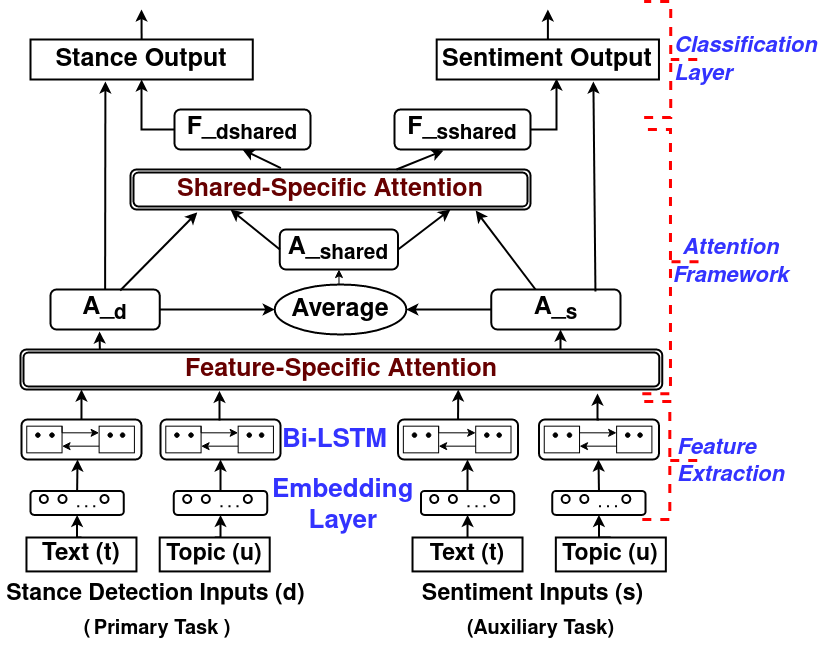}
  \caption{Architecture of the Shared-Private Multi-Task \\  (SP-MT) Framework }
  \label{sp_mt}
  \vspace{-0.5cm}
\end{minipage}
\begin{minipage}{.5\linewidth}
\includegraphics[width=1.05\linewidth]{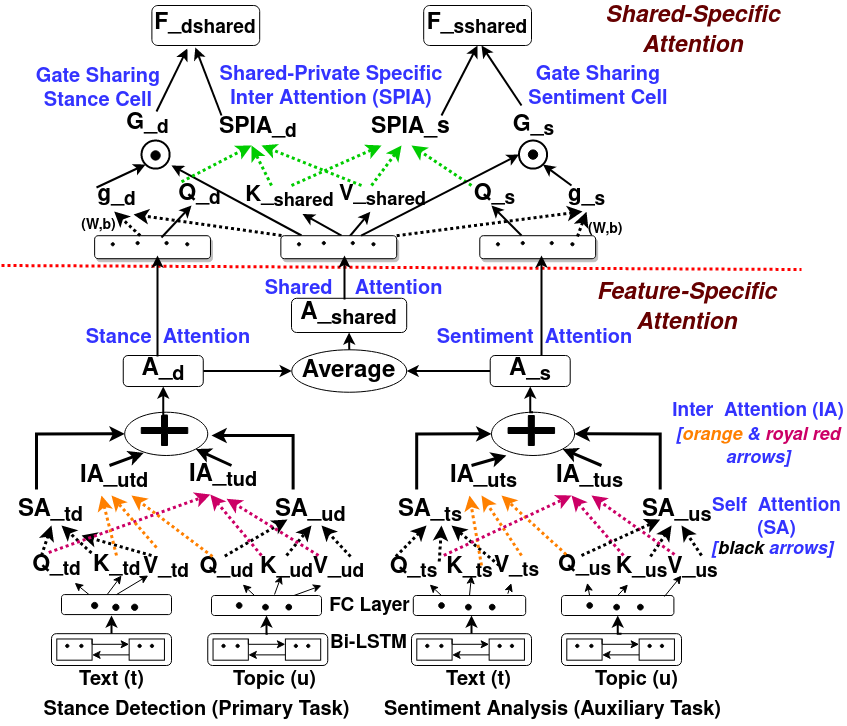}
  \caption{Attention Framework : (Top) Shared-Specific Attention; (Bottom) Feature-Specific Attention}
  \label{attn_frame}
  \vspace{-0.4cm}
\end{minipage}
\end{figure*}

\noindent(ii)\textit{\textbf{Shared-Private Multi-Task Model (SP-MT)}}: In the model SP-MT, we have two separate feature spaces for the two tasks that capture task-specific features, and a single shared feature space that captures the task-invariant features (refer Fig. \ref{sp_mt}). The final features are the concatenation of the features from the private space and the shared space, which are then fed into the classification layer to generate the output for both tasks.
\par\noindent The input and output of each model component for both variants (Figures \ref{shared_only} and \ref{sp_mt}) are mentioned in the following subsection \ref{modelComp}. We now describe each of the model components in detail.

\subsection{Components of the Model}\label{modelComp}
\subsubsection{Feature Extraction\\} 
\par\noindent {\textbf{Text}} : Each tweet text `T' contains $n_t$ number of words, where the embedding of each word $w_1,. . ., w_{n_{t}}$ is acquired from BERT \cite{DBLP:conf/naacl/DevlinCLT19} with dimension(d) = 768. We obtain final embedding for each tweet text as $T \in \mathbb{R}^{n_{t} \times d} $. Since Bi-LSTM has shown excellent performance in text classification due to its ability to learn long-term dependencies and incorporate past and future context information without retaining duplicate information \cite{DBLP:conf/coling/ZhouQZXBX16}, we use Bi-LSTM to sequentially encode the embedded input text representations. The embedded text is fed to Bi-LSTM with dimension $d_l$, which learns the long-term context dependent semantic features into hidden states. The final hidden matrix of text is $H_t \in \mathbb{R}^{n_{t} \times 2d_l} $. 
\par\noindent {\textbf{Topic}} : Each tweet is tagged with the number of m-most similar topics based on the probability score assigned to each of the topics, where each topic is represented by the top `p' topic words created using BERTopic (described in section \ref{data_preprocess}). Here we represent the topic feature as `U' containing a set of $n_u$ words ($n_u$ $\le m\times p$), where the representation of each word $w_1,..., w_{n_{u}}$ are obtained from BERT with $d=768$. We obtain final embedding for each tweet topic as $U \in \mathbb{R}^{n_{u} \times d} $. This representation of topic is then fed to the Bi-LSTM layer with $d_l$ that sequentially encodes these representation, and gives the final hidden matrix of topic $H_u \in \mathbb{R}^{n_{u} \times 2d_l} $.

\subsubsection{Attention Framework} \label{attn_section}
Attention mechanism has been used as an important component across a wide range of NLP models \cite{bahdanau2014neural}. Typically, the attention layer concentrates on the relevant part of the input and extracts the most important information from the input. We apply the attention framework similar to \cite{vaswani2017attention}, in which the authors consider an attention function as a mapping to a set of queries, keys, and values. To obtain queries, keys, and values for the final feature representations, we pass the hidden matrix output from the Bi-LSTM layer of text ($H_t$) and topic ($H_u$), respectively, through three fully connected layers of dimension $d_a$. There are two triplets of query, key, and value for text ($Q_t,K_t,V_t$) and topic ($Q_u,K_u,V_u$) in the model SO-MT, while we have a total of four triplets for the model SP-MT, forming two pairs of two triplets each for text and topic, which are used for stance detection (($Q_{td},K_{td},V_{td}$),($Q_{ud},K_{ud},V_{ud}$)) and sentiment (($Q_{ts},K_{ts},V_{ts}$),($Q_{us},K_{us},V_{us}$)) tasks respectively. Fig. \ref{attn_frame} visually shows the two attention frameworks used in our model: \textit{feature-specific attention} and \textit{shared-specific attention}. The lower part of the figure shows how different queries, keys, and values are encoded to obtain self-attention and inter-attention, which form the two sub-modules of feature-specific attention. The upper part of the figure shows the connections between queries, keys, and values to achieve shared-specific attention. In the following, we describe in detail the attention mechanisms used in our study.
\par \noindent \textbf{Feature-Specific Attention}\label{mod_section} We apply two types of attention to the features to capture the most informative parts of them. Fig. \ref{attn_frame} (bottom) shows the visual representation and connections of feature-specific attention. Feature-specific attention is further divided into Self Attention (SA) and Inter Attention (IA). 

\par \noindent \textbf{\textit{Self Attention (SA)}} We use Self Attention (SA) to relate different positions of a single sequence of say tweet text or topic to quantify the most important part of that sequence \cite{vaswani2017attention}.  
\begin{gather}\label{eqn1}
        SA_j = softmax(Q_{j}K^T_{j})V_{j}
\end{gather}
\par \noindent SA scores are calculated using the equation \ref{eqn1}, where $SA_t \in \mathbb{R}^{n_{t} \times d_a} $, and $SA_u \in \mathbb{R}^{n_{u} \times d_a} $. Here, two SA scores are computed for SO-MT, while four such SA scores are required for SP-MT model ($SA_{td}, SA_{ud}, SA_{ts},SA_{us}$) (as shown with black dotted arrow connections in Fig. \ref{attn_frame} (bottom).

\par \noindent \textbf{\textit{Inter Attention (IA)}} We find out the Inter Attention (IA) scores to learn the interdependence between different features. IA scores are determined using below equations where query of one feature is intervened with key and value of the other. IA scores help to reveal the significant contributions amongst different inputs to learn optimal features for both tasks. The equations that represent the IA scores for text and topic ($IA_{tu}$, $IA_{ut}$) are:
\begin{gather}
    IA_{tu} = softmax(Q_{t}K^T_{u})V_{u}, \label{ia_1} \\
    IA_{ut} = softmax(Q_{u}K^T_{t})V_{t}, \label{ia_2}
\end{gather}
where $IA_{tu} \in \mathbb{R}^{n_{t} \times d_a} $, and $IA_{ut} \in \mathbb{R}^{n_{u} \times d_a} $. IA equations are represented graphically with orange and royal red dotted arrows in Fig. \ref{attn_frame} (bottom) part. The SA and IA scores are then concatenated finally, where A is directly used for shared-only (SO) variant (refer Fig. \ref{shared_only}) while average of attention vector ($A_{shared}$) specific to stance and sentiment tasks ($A_d$,$A_s$) is used for Shared-private (SP) variant of model (mentioned in Fig. \ref{attn_frame} (bottom)). 
\begin{gather}
    A    = concat(SA_{t},SA_u, IA_{tu},IA_{ut}), \\
    A_{d} = concat(SA_{td},SA_{ud}, IA_{tud},IA_{utd})), \label{attn_pol}\\
    A_{s} = concat(SA_{ts},SA_{us}, IA_{tus},IA_{uts})),\\
    A_{shared} = Average(A_{d},A_{s}) \label{attn_sh}
\end{gather}

\par \noindent \textbf{Shared-Specific Attention}\label{sharedattn_section} Some of the works mentioned in section \ref{rw_sent} focused on the orthogonal relationship between sentiment and stance detection. Although climate change deniers and proponents dominate in sentiment, there are a few examples where sentiment does not match attitude/stance (see Table \ref{tab:data1}). \cite{wu2019different} also mentioned the disadvantage of the shared-private model of multi-task learning, explaining that the shared space usually mixes some task-relevant features, which makes learning different tasks difficult. 
To solve the above problems, we use the \cite{wu2019different} inspired shared-specific attention, which filters out the useless features that interfere with the model prediction and only pay attention to the selected features from the shared layer that lead to the correct predictions of the SP-MT model (see Fig. \ref{sp_mt}). Next, we describe the sub-modules to achieve the desired result (as shown in Fig. \ref{attn_frame} (top))

\par \noindent \textbf{\textit{Gate Sharing Cell}} We use similar approach used by authors \cite{wu2019different} where a single gate mechanism removes the useless shared features from shared layer. We first express the cell with reference to stance detection task. The stance specific, and shared attention scores ($A_d, A_{shared})$ from equations (\ref{attn_pol},\ref{attn_sh}) are passed through dense layers with $d_s$ units. The weights and biases are captured when passing $A_d$ through dense layer which are used for $A_{shared}$, and is expressed as gated sharing cell. 
\begin{eqnarray}
     g_d = \sigma(W_d . A_{shared}  + b_d)
    \label{eq:lstm}
\end{eqnarray}
where, $W_{d} \in \mathbb{R}^{n_{t}(d_s) \times n_{t}(d_s)} $ and $b_{d} \in \mathbb{R}^{1 \times n_{t}(d_s)} $. Similar equations are followed for sentiment task also, hence, we do not iterate here.
The final output of the shared features for both the task after filtering will be represented as :
\begin{gather}
    G_d    = g_d \odot A_{shared}, \\
    G_s    = g_s \odot A_{shared}
\end{gather}
where $\odot$ denotes element-wise multiplication. Fig. \ref{attn_frame} (top) shows the connections of gate sharing cell for stance detection and sentiment tasks.
\par \noindent \textbf{\textit{Shared-Private specific Inter Attention (SPIA)}} We use similar concept of the inter attention of feature-specific attention (equations \ref{ia_1}, \ref{ia_2}). We capture the important shared features relevant to the specific task, by using query matrix of the particular task (stance/sentiment) and keys and values of the shared task. The attention vectors ($A_d,A_s$) are passed through fully connected layers with $d_s$ units to create 
$Q_d,Q_s$ for stance and sentiment tasks, while $A_{shared}$ is passed through dense layer to generate $K_{shared}$, and $V_{shared}$. 
The equations are represented visually with green dotted arrows in Fig. \ref{attn_frame} (top) part.
\begin{gather}
    SPIA_{d} = softmax(Q_{d}K^T_{shared})V_{shared}, \\
    SPIA_{s} = softmax(Q_{s}K^T_{shared})V_{shared}
\end{gather}

\par \noindent \textbf{\textit{Fusion}} The final output of the shared layer is the fusion of the output of the gated cell and shared-private specific inter attention. Recently, fusion technique with absolute difference and element-wise product is found to be effective in \cite{mou2015natural}.
\begin{gather}
    C_1=[G_d;SPIA_d;G_d - SPIA_d;G_d \odot SSIA_d];\\
    C_2=[G_s;SPIA_s;G_s - SPIA_s;G_s \odot SSIA_s];\\
    F_{dshared} = \tanh(W_{fd}.C_1  + b_{fd}), \\
    F_{sshared} = \tanh(W_{fs}.C_2  + b_{fs})
\end{gather}

\subsubsection{Classification Layer} 
The final representation of the tweet obtained($A_d$,$A_s$) is passed through separate outputs for stance and sentiment tasks (for SO-MT model (Fig. \ref{shared_only}) ), however individual task specific tweet representations along with the shared layer representations are passed through two output channels, subjected to polarisation ($A_d$,$F_{dshared}$) and sentiment ($A_s$,$F_{sshared}$) tasks for SP-MT Model (Fig. \ref{sp_mt}). The task specific and shared loss are used as 
\begin{gather}
     L_{total} = L_{task} + \lambda L_{shared}
\end{gather}
where $\lambda$ = $0.5$ is a hyper-parameter \cite{DBLP:conf/acl/LiuQH17}.

\section{Experiment}
\subsection{Datasets} We evaluate our model performance and compare with the other baselines on the two datasets:

\par\noindent\textbf{Climate Change Data} The details of the data collection and statistics are covered in section \ref{dataset} and  table \ref{tab:data}.
\par\noindent\textbf{SemEval} is provided in the SemEval-2016 shared task 6.A on tweet stance detection \cite{mohammad2016semeval}. Each tweet is in favor, against or neutral corresponding to  one of the five targets: \textit{Atheism, Climate Change is a Real Concern, Feminist Movement, Hillary Clinton, and Legalization of Abortion(Abortion)}. There has been several works that use this benchmark dataset for stance classification.
\begin{table*}[]
\centering
\scalebox{0.75}{
\begin{tabular}{|c|c|c|c|c|c|c|c|c|}
\hline
\multirow{2}{*}{\textbf{Model}} &\multicolumn{4}{c|}{\textbf{Single-Task Stance Detection}} &\multicolumn{4}{c|}{\textbf{Single-Task Sentiment}}\\ 
\cline{2-9}
 &\multicolumn{2}{c|}{\textbf{Text}} &\multicolumn{2}{c|}{\textbf{Text+Topic}} &\multicolumn{2}{c|}{\textbf{Text}} &\multicolumn{2}{c|}{\textbf{Text+Topic}} \\ \cline{2-9} 
 &\textbf{Acc.} & \textbf{F1} & \textbf{Acc.} & \textbf{F1} & \textbf{Acc.} & \textbf{F1} & \textbf{Acc.} & \textbf{F1}\\ \hline
Shared-only (SO) &76.73$\pm$2.48 &70.75$\pm$2.11 &79.56$\pm$1.10 &75.72 $\pm$1.08 &71.61$\pm$0.45 &70.11$\pm$0.79 &75.82$\pm$2.03 &74.63$\pm$2.41\\ \hline
SO + Self Attn. (SA)  &80.29$\pm$0.55 &74.24$\pm$0.31 &82.88$\pm$0.78 &77.56 $\pm$0.87 &73.49$\pm$0.92 &72.57$\pm$1.27 &77.68$\pm$1.27 &76.59$\pm$1.66\\ \hline
   SO + Inter Attn. (IA)&81.33$\pm$1.04 &73.35$\pm$1.41 &83.13$\pm$1.42 &76.04 $\pm$1.11 &73.68$\pm$3.20 &72.23$\pm$3.14 &77.93$\pm$2.01 &74.11$\pm$1.98\\ \hline
{\begin{tabular}[c]{@{}l@{}}   SO + SA + IA \\ (Feature-Specific Attn.)\end{tabular}} &82.15$\pm$0.05 &76.78$\pm$0.81 &\textbf{85.01$\pm$1.05} &\textbf{80.03 $\pm$2.42} &76.21$\pm$0.84 &74.95$\pm$1.02 &\textbf{80.81$\pm$1.29} &\textbf{79.28$\pm$1.44}\\ \hline
\end{tabular}
}
\caption{Results of the Single-Task Stance Detection and Sentiment tasks for various combinations}
\label{single-task}
\end{table*}
\begin{table*}[]

\centering
\scalebox{0.75}{
\begin{tabular}{|c|c|c|c|c|c|c|c|c|}
\hline
\multirow{4}{*}{\textbf{Model}} &\multicolumn{8}{c|}{\textbf{Multi-Task Stance Detection + Sentiment}}\\ 
\cline{2-9}
&\multicolumn{4}{c|}{\textbf{Stance Detection}} &\multicolumn{4}{c|}{\textbf{Sentiment}}\\ 
\cline{2-9}
 &\multicolumn{2}{c|}{\textbf{Text}} &\multicolumn{2}{c|}{\textbf{Text+Topic}} &\multicolumn{2}{c|}{\textbf{Text}} &\multicolumn{2}{c|}{\textbf{Text+Topic}} \\ \cline{2-9} 
 &\textbf{Acc.} & \textbf{F1} & \textbf{Acc.} & \textbf{F1} & \textbf{Acc.} & \textbf{F1} & \textbf{Acc.} & \textbf{F1}\\ \hline
Shared-only (SO) &81.16$\pm$1.89 &77.01$\pm$2.61 &84.86$\pm$1.13 &81.66 $\pm$3.28 &75.44$\pm$1.23 &74.40$\pm$1.75 &80.49$\pm$2.05 &79.19$\pm$2.8\\ \hline
Shared-Private (SP)&84.64$\pm$1.95 &81.53$\pm$2.55 &87.47$\pm$2.01 &84.24 $\pm$2.04 &77.93$\pm$2.09 &74.11$\pm$2.19 &85.28$\pm$0.91 &84.58$\pm$1.02\\ \hline
{\begin{tabular}[c]{@{}l@{}}Shared-Private (SP) +  \\ Feature-Specific Attn.\end{tabular}}  &86.49$\pm$2.29 &81.67$\pm$3.21 &91.31$\pm$1.06 &85.93 $\pm$1.22 &81.46$\pm$0.71 &80.71$\pm$0.39 &85.46$\pm$0.9 &84.73$\pm$1.10\\ \hline
{\begin{tabular}[c]{@{}l@{}}Shared-Private (SP) +  \\ Shared-Specific Attn.\end{tabular}}  &88.10$\pm$1.04 &84.09$\pm$1.39 &92.29$\pm$1.32 &86.12 $\pm$1.4 &81.67$\pm$0.61 &80.89$\pm$0.32 &87.59$\pm$0.45 &87.07$\pm$0.39\\ \hline
{\begin{tabular}[c]{@{}l@{}}Shared-Private +  \\ Feature-Sp. Attn. + \\ Shared-Sp. Attn. \\ (SP-MT) \end{tabular}}  &89.99$\pm$2.67 &86$\pm$2.02 &\textbf{93.95$\pm$1.27} &\textbf{90.24 $\pm$1.16} &84.60$\pm$0.44 &83.98$\pm$0.81 &\textbf{89.08$\pm$1.01} &\textbf{88.48$\pm$1.60}\\ \hline

\end{tabular}
}
\caption{Results of the Multi-Task Stance Detection and Sentiment tasks for various combinations}
\label{multi-task}
\end{table*}
\subsection{Set-up} We use the python-based library Keras\footnote{https://keras.io/} at various stages of our implementations. 
For the experiments, we perform stratified k-fold cross-validation on our dataset, oversample the minority class (deniers) in the k-1 training data using the sklearn resampling technique, and report the averaged scores and standard deviation (over 5 folds) for the accuracy and F1 scores. We select $m=5$ and $p=10$, which fits our dataset well, where $m$ denotes the number of most similar topics and each topic contains $p$ number of words for the topic feature of each tweet. In the feature extraction sub-module, Bi-LSTM ( $d_l$ ) with $100$ memory cells is used. The dimensions $d_a$ and $d_s$ of the fully connected layers used in the attention framework to extract queries, keys and values for feature-specific and shared-specific attention (refer section \ref{attn_section}) are used with $100$ units each. The stance and sentiment output channels contain $2$ and $3$ output neurons, respectively. The loss functions binary cross-entropy and categorical cross-entropy are used for the stance and sentiment output channels, respectively. The experiments are run on an NVIDIA GeForce GTX 1080Ti GPU and the models are optimized using Adam optimizer with a learning rate of $0.0001$. All these values are selected using TPE in the hyperopt python library \cite{bergstra2013hyperopt} and after a thorough sensitivity analysis of the parameters that minimise the loss functions.

\subsection{Baseline Techniques} \label{baselines_section}
We compare our proposed approach with the following baselines on our climate change dataset :
\par\noindent\textbf{Logistic regression \cite{argyris2021using}}: This study use logistic regression with Count Vectorizer feature extraction method to classify vaccine-related tweets into provaccine, antivaccine, and neutral stances.

\par\noindent\textbf{ESD \cite{vychegzhanin2021new}}: The authors form a relevant feature set using an ensemble of feature selection methods and propose the model ESD by selecting an optimal ensemble of classifiers. They evaluate the performance of the model using the UKP Sentential Argument Mining Corpus and the SemEval-2016 dataset.

\par\noindent\textbf{HAN \cite{wang2020neural}}: In this article, researchers proposed a hierarchical attention neural model, focusing on different features such as document, sentiment, dependency, and argument representations. They evaluated the model performance on SemEval-2016 and the H\&N14 dataset.

\par\noindent\textbf{AT-JSS-LEX \cite{li2019multi}}: is a multi-task framework for stance detection with sentiment analysis as auxiliary task. The attention mechanism of the model is guided by target-specific attention along with sentiment and stance lexicons. 

\par\noindent\textbf{MNB \cite{kabaghe66classifying}}: Multinomial naive bayes performed better with respect to other models proposed in the study, to classify tweets into positive, negative or neutral beliefs towards climate change.

\par\noindent\textbf{DNN \cite{chen2019detecting}}: Deep Neural Network (DNN) is used as a classifier to identify users who either believe or deny climate change based on the content of tweets. The model's performance is assessed on the real-time collection of climate change twitter data.
\par\noindent\textbf{SVM-ngram \cite{sobhani2016detecting}}: is trained on word and character n-grams features for stance detection task on SemEval 2016 dataset. The model surpassed the best model in SemEval-2016 competition.

\par\noindent We evaluate our model performance on \textit{SemEval 2016} dataset and contrast with the these state-of-the art models :  \textbf{ESD} \cite{vychegzhanin2021new}, \textbf{HAN} \cite{wang2020neural}, \textbf{AT-JSS-LEX} \cite{li2019multi}, and \textbf{SVM-ngram} \cite{sobhani2016detecting} as described above.

\section{Result and Analysis}
In this section, we investigate the performance of the proposed approach. We first compare different single-task and multi-task variants and then compare them with the state-of-the-art methods mentioned in section \ref{baselines_section}. We also analyze the importance of each feature and the different variants of the attention framework. We report all the results of the five-fold cross-validation (mean and standard deviation of accuracy and F1 score) for the different combinations of the proposed system.
\begin{table*}
\centering
\scalebox{0.85}{
\begin{tabular}{|l|l|l|l|l|l|}
\hline
\textbf{No.}&\textbf{Tweet} &\textbf{Sentiment} & \textbf{True} & \textbf{{\begin{tabular}[c]{@{}l@{}}Predicted \\ Stance \end{tabular}}} & \textbf{{\begin{tabular}[c]{@{}l@{}}Predicted \\ Stance + \\ Sentiment \end{tabular}}} \\ \hline
1. &My family support Oil and Gas! ClimateHoax &positive  &denier &\color{red}believer &denier \\ \hline
2. &{\begin{tabular}[c]{@{}l@{}}Once again brainwashing kids to push the green tax agenda,\\ under the globalwarminghoax umbrella! Stop\end{tabular}} &negative  &denier &denier &denier \\ \hline
3. &His Green BS policies will send us back to the Dark Ages. ClimateChangeHoax &negative  &denier &\color{red}believer &denier \\ \hline
4. &{\begin{tabular}[c]{@{}l@{}}ClimateHoax The climate has fluctuated since the  time of creation,\\ and nothing those people will do can change that one way or the other\end{tabular}} &neutral  &denier &\color{red}believer &denier \\ \hline
5. &{\begin{tabular}[c]{@{}l@{}}And yet, there are those who deny climate change?? Ice-shelves breaking off,\\ heat waves, etc. Sad ScienceIsReal\end{tabular}} &negative  &believer &\color{red}denier &believer \\ \hline
6. &{\begin{tabular}[c]{@{}l@{}}Have you seen this? Its very moving. We definitely need more ClimateAction\end{tabular}} &positive  &believer &believer &believer \\ \hline
7. &{\begin{tabular}[c]{@{}l@{}}For those adamant that global warming is real, THIS is Today in Alaska.\\ Four inches of snow overnight, and  still coming down! ClimateChange\end{tabular}} &neutral  &denier &\color{red}believer &\color{red}believer \\ \hline


8. &{\begin{tabular}[c]{@{}l@{}}I am glad you went by plane. Way better for our climate instead of zoommeetings...\end{tabular}} &positive &believer &\color{red}denier &\color{red}denier \\ \hline


9. &{\begin{tabular}[c]{@{}l@{}}Over 60° today. Over 6" of snow tomorrow. But yeah, climatechange is total bullshit right?\end{tabular}} &negative &believer &\color{red}denier &\color{red}denier \\ \hline


\end{tabular}

}
 \setlength{\abovecaptionskip}{2pt}
\caption{Few example tweets with ground truth and predicted labels for single and multi-task models}
\label{examples}
\end{table*}
The Tables \ref{single-task} and \ref{multi-task} illustrate the results of the single-task and the various combinations of the proposed multi-task models for both the stance detection and sentiment tasks. It is evident that the addition of topic words consistently improves the performance of the models. This improvement means that the proposed architecture makes very effective use of the interaction between input features. This shows the importance of incorporating multiple features for various analysis tasks.

\subsection{Comparison amongst Single-Task and Multi-task Framework}
\par \noindent From the tables \ref{single-task} and \ref{multi-task}, the multi-task variants perform better than the single-task variants by achieving an average macro F1 score of $90.24$ and $88.48$ for the stance detection (primary) and sentiment analysis (auxiliary) tasks respectively. The results show that the sentiment and stance tasks improve each other's performance when learned together. The single stance detection task is able to correctly label some tweets from deniers and believer that contain predominantly negative and positive sentiments, respectively (examples $2$ and $6$ from Table \ref{examples}). However, examples $1$, $4$, and $5$ from Table \ref{examples} clearly show that the stance task, together with the sentiment analysis task, is able to unambiguously identify denier and believers tweets with the corresponding less dominant positive, neutral, and negative sentiment polarities. As stated earlier, we consider sentiment analysis as an auxiliary task that supports the main task, i.e., stance detection. However, we report the performance of the sentiment task for the proposed model for both single-task and multi-task frameworks in the tables \ref{single-task} and \ref{multi-task} to illustrate the impact of the main task on the auxiliary task and to show that the multiple features in the form of tweet text and topic words, as well as the attention framework, also benefit the sentiment classification task. However, we do not make explicit efforts to improve the model performance on the auxiliary task.

\begin{table}
\centering
\scalebox{0.70}{
\begin{tabular}{|l|l|l|}
\hline
\textbf{Model} &\textbf{Training Time (secs)} &\textbf{Mean Accuracy}  \\ \hline
{\begin{tabular}[c]{@{}l@{}} Single Task Best \\ (SO + SA + IA) [Table 5]\end{tabular}} &870 &85.01 \\ \cline{1-3}
\multicolumn{3}{|c|}{\textbf{Multi Task Variants} [Refer Table 6]} \\ \hline
Shared-only (SO) &918 &84.86 \\ \hline
Shared-Private (SP) &1218 &87.47 \\ \hline
{\begin{tabular}[c]{@{}l@{}}Shared-Private (SP) +  \\ Feature-Specific Attn.\end{tabular}} &1419 &91.31 \\ \hline
{\begin{tabular}[c]{@{}l@{}}Shared-Private (SP) +  \\ Shared-Specific Attn.\end{tabular}} &1506 &92.29 \\ \hline
{\begin{tabular}[c]{@{}l@{}}Shared-Private +  \\ Feature-Sp. Attn. + \\ Shared-Sp. Attn. (SP-MT) \end{tabular}} &1791 &93.95 \\ \hline
\end{tabular}
}
\caption{Training time of Different Text + Topic Models}
 \setlength{\abovecaptionskip}{2pt}
\label{tab_training}
\end{table}

\subsection{Comparison amongst Different Multi-task Frameworks} 
Table \ref{multi-task} shows the improvement of the multi-task framework from the shared-only variant (Fig. \ref{shared_only}) to the shared-private multi-task model(SP-MT) (Fig. \ref{sp_mt}). The inclusion of feature-specific and shared-specific attention frameworks helped the multi-task models focus on the important parts of the features and effectively discard the useless shared features, resulting in a $7.40\%$ increase in accuracy and a $12.46\%$ increase in F1 score. Furthermore, in Table \ref{tab_training} we give the training times of the best performing single-task model and different variants of the multi-task model for 20 epochs to analyze the additional time required by the best performing multi-task model with text and topic as input features (SP-MT) compared to other variants. As can be seen from the table \ref{tab_training}, SP-MT requires about $15$ more minutes (approximately twice the time) to achieve a $10.51$\% improvement in accuracy compared to the best performing single-task model, while SP-MT requires $9.5$ more minutes to achieve a $7.4$\% improvement in performance compared to the shared-private multi-task variant.
\par \noindent All results reported here are statistically significant as we performed a t-test at the $5\%$ significance level \cite{welch1947generalization} against the null hypothesis, which states that the mean accuracy/F1 score of all the multi-task variants is more when compared to the the best performing proposed model SP-MT (Shared-Private + Feature-Specific Attention +Shared-Specific Attention) [refer table \ref{multi-task}]. If the p-value is significant ($p<0.05$), we reject the null hypothesis. Our best performing proposed model outperforms all the other multi-task variants while meeting statistical significance under t-tests ($p<0.05$). For the confidence analysis, we also report the p-values and t-test statistics of all the multi-task variant models compared to the best performing model in tables $1$ and $2$ in Supplementary.
\subsection{Comparison with the Baseline Methods}
In Table \ref{baselines}, we report the results for the baseline methods by re-implementing them on the Climate Change dataset (section \ref{dataset}). It is observed that our proposed multi-task approach SP-MT outperforms the SOTA approaches in terms of accuracy and F1 score. Our best performing model achieved better results compared to ESD \cite{vychegzhanin2021new} and HAN \cite{wang2020neural}. This highlights that the shared-private multi-tasking approach takes advantage of task-specific and invariant features to improve classification task performance. Although the AT-JSS-LEX \cite{li2019multi} model was implemented with a multi-tasking approach, our model performs better because it keeps the task-dependent and task-independent feature spaces separate and removes the useless shared features that hinder task performance of the stance detection, demonstrating the importance of the shared-specific attention framework. It is also observed that the methods that use sentiment features (ESD, HAN, and AT-JSS-LEX) perform better than the other baselines. This proves the importance of the proposed sentiment analysis approach for climate change. Our best performing single-task polarization framework (Table \ref{single-task}) also outperforms MNB, DNN, and SVM-ngram approaches. This justifies the benefits of using topic words in addition to tweet text and feature-specific attention framework to improve the performance of the model. \\
\noindent Consequently, we also performed a comparative analysis of the proposed multi-tasking approach SP-MT with the state-of-the-art models (SOTA) on the SemEval 2016 dataset. The model is trained with three polarized classes \textit{(Favour, Against, None)} and the metrics ($F_{avg}$, $MacF_{avg}$) are evaluated according to the procedure defined in \cite{li2019multi}. Table \ref{sem_eval} shows that our approach outperforms other methods with an overall $MacF_{avg}$ value of $66.84$. Our proposed framework performs better in the climate, feminism, and abortion domains, while the $F_{avg}$ values are comparable in the atheism and Hillary domains, showing that our framework generalises well in different domains.

\subsection{{Error Analysis}} \label{Error_analysis}
We perform an in-depth error analysis to understand where the proposed model has faltered. These are the following scenarios: \textbf{(i)} The climate change dataset is an imbalanced dataset with a high proportion of believer tweets, resulting in low F1 values compared to accuracy. Although we applied oversampling to partially counter this problem, even finer categories of believers can be identified and labeled, which can be beneficial for the model to learn different classes with a clear separation of distribution in tweets, such as ``tweet conveys causes of climate change'' or ``tweet believes in human-caused climate change''. \textbf{(ii)} We determine the frequency of unigrams and bigrams extracted using TF-IDF and find that some of the denier's tweets containing either rarely used keywords or keywords frequently used in believers' tweets were misclassified. For example, the denier's tweet in example $7$ (table 7) contains words such as \textit{real, snow overnight}, which are most commonly found in believers' tweets and confuse the model and lead to an incorrect prediction. \textbf{(iii)} We investigated that of the total misclassified denier tweets, $35.7\%$ of the tweets contained sarcasm to express their denial. Of the sarcastic denier tweets, $50.16\%$ of the tweets have positive sentiment, $31.70\%$ have neutral sentiment, and the rest have negative sentiment, while $25.78\%$ of the misclassified believer tweets have sarcastic labels (examples $8$ and $9$ of table 7). The labeling of sarcasm is based on the majority vote of three trained annotators with an inter-rater agreement of $0.78$, calculated with the Fleiss-Kappa measure. This motivated us to investigate the presence of sarcasm in climate change tweets to further improve the performance of the model as a part of our future work. 
\begin{table}
\centering
\scalebox{0.67}{
\begin{tabular}{|c|c|c|}
\hline
\textbf{Model} &\textbf{Accuracy} &\textbf{F1-score} \\ \hline
Proposed SP-MT (Fig. \ref{sp_mt}) &\textbf{93.95} &\textbf{90.24} \\ \hline
LR (Argyris et al. 2021) &81.48 &81.00 \\ \hline
ESD \cite{vychegzhanin2021new} &89.65 &85.11 \\ \hline
HAN \cite{wang2020neural} &89.47 &86.00 \\ \hline
AT-JSS-LEX \cite{li2019multi} &88.02 &84.01 \\ \hline
MNB \cite{kabaghe66classifying} &85.44 &78.08 \\ \hline
DNN \cite{chen2019detecting} &84.61 &76.23 \\ \hline
SVM-ngram \cite{sobhani2016detecting} &85.55 &66.33 \\ \hline
\end{tabular}

}
 \setlength{\abovecaptionskip}{2pt}
  \setlength{\belowcaptionskip}{-6pt}
\caption{Results of Proposed Framework SP-MT with baselines on our Climate Change Dataset}
\label{baselines}
\end{table}

\begin{table}
\centering
\scalebox{0.61}{
\begin{tabular}{|c|c|c|c|c|c|c|}
\hline
\textbf{Model} &\textbf{{\begin{tabular}[c]{@{}l@{}}Atheism\\ $F_{avg}$ \end{tabular}}} &\textbf{{\begin{tabular}[c]{@{}l@{}}Climate\\ $F_{avg}$ \end{tabular}}} &\textbf{{\begin{tabular}[c]{@{}l@{}}Feminism\\ $F_{avg}$ \end{tabular}}} &\textbf{{\begin{tabular}[c]{@{}l@{}}Hillary\\ $F_{avg}$ \end{tabular}}} &\textbf{{\begin{tabular}[c]{@{}l@{}}Abortion\\ $F_{avg}$ \end{tabular}}} &Mac $F_{avg}$ \\ \hline
Proposed SP-MT(Fig.\ref{sp_mt}) &69.5 &\textbf{63.5} &\textbf{63.2} &67.5 &\textbf{70.5} &\textbf{66.84} \\ \hline
{{\begin{tabular}[c]{@{}l@{}}ESD \\ (Vychegzhanin and\\ Kotelnikov 2021) \end{tabular}}} &66.64 &43.82 &62.85 &67.79 &64.94 &61.20 \\ \hline
HAN \cite{wang2020neural} &\textbf{70.53} &49.56 &57.50 &61.23 &66.16 &61.00 \\ \hline
{{\begin{tabular}[c]{@{}l@{}}AT-JSS-LEX\\ \cite{li2019multi}\end{tabular}}} &69.22 &59.18 &61.49 &\textbf{68.33} &68.41 &65.33 \\ \hline
{{\begin{tabular}[c]{@{}l@{}}SVM-ngram\\(Sobhani,  Mohammad,\\  and  Kiritchenko2016) \end{tabular}}}  &65.19 &42.35 &57.46 &58.63 &66.42 &58.01 \\ \hline
\end{tabular}
}
 \setlength{\abovecaptionskip}{2pt}
\caption{Results of Proposed Framework SP-MT with baselines on SemEval 2016 Dataset}
\label{sem_eval}
\end{table}

\section{Privacy and Ethics}
Although social media offers innovative ways to raise awareness about climate change, phenomena such as climate denial and climate delay have become a serious problem for scientists and the government to convince people of the importance of understanding the current climate crisis. Climate change deniers are not only skeptical about climate change but also emphasize the disadvantages of all measures proposed to combat climate change and abandon the idea that it is not possible to prevent climate change. This often leads to the spread of misinformation, resulting in a delay in the implementation of effective climate change mitigation measures \cite{zhou2021confirmation}. Since our work is dedicated to classifying Twitter content into climate change deniers or believers, the proposed approach can be useful for government agencies, researchers, and tech companies that monitor such content on social media to identify and intervene tweets from climate change deniers. The proposed approach is useful in combating climate misinformation by identifying posts by climate change deniers and reducing the spread of such content that is deemed false or misleading.
\par \noindent The input feature of the proposed model, such as the tweet text, is available as soon as the user posts something. However, the topic feature can be extracted by performing topic modeling for a collection of tweets after a fixed interval, e.g., after every 5, 10, 15, or t minutes of duration. Therefore, our proposed approach can be used in a real-time environment by interested agencies and authorities to classify social media content into one of the two polarized classes.
\par \noindent Although we conduct our work with public data from social media, however, we are committed to protecting the privacy of individuals and therefore avoid providing personally identifiable content. The dataset that is made publicly available consists only of tweet IDs and comments.

\section{{Conclusion and Future Work}}
In this paper, we focus on the importance of classifying Twitter content into climate change deniers and believers, because climate change deniers emphasize the downsides of any action to address climate change, which often leads to the spread of misinformation and delays the implementation of effective action to mitigate climate change. Our proposed approach, implemented in real-time, can be useful for government agencies, researchers, and tech companies in combating climate misinformation by identifying content that denies climate change and reducing the spread of such posts that are considered misleading. \\
\noindent In this work, we investigated the role of sentiment in classifying stance of the tweets related to climate change. We curate a novel dataset that includes annotations for both stance detection and sentiment analysis tasks, which will be useful to the research community in exploring other needed classification tasks. We propose a shared-private multi-task framework for the optimization of stance detection task benefiting from the sentiment analysis (auxiliary task). The proposed module uses feature-specific and shared-specific attention to fuse multiple features and learn useful and relevant private and shared features for both tasks. The results show that multi-tasking increased the performance of the stance detection task compared to its uni-modal and single-task variants. Although we examined the performance of the proposed approach in detecting attitudes in the domain of climate change, the performance of the model on the SemEval dataset shows that it is much more broadly applicable beyond the domain of climate change, suggesting that our framework can generalize well in different domains.
Future work will attempt to analyse what other aspects of natural language, such as sarcasm, aspect-based sentiment, and emotion recognition, might help to more accurately classify  attitudes toward climate change. The inclusion of other modality encodings such as images, emoji, and advanced architectures will also be the subject of our future work.


\bibliographystyle{aaai}
\bibliography{bibliography.bib}

\end{document}